# Computationally Efficient Target Classification in Multispectral Image Data with Deep Neural Networks

Lukas Cavigelli[a,*], Dominic Bernath[a], Michele Magno[a,b], Luca Benini[a,b]

[a]ETH Zurich, Integrated Systems Laboratory, Gloriastr. 35, CH-8092 Zurich, Switzerland
[b]University of Bologna, DEI, Viale Risorgimento 2, I-40126 Bologna, Italy

## ABSTRACT

Detecting and classifying targets in video streams from surveillance cameras is a cumbersome, error-prone and expensive task. Often, the incurred costs are prohibitive for real-time monitoring. This leads to data being stored locally or transmitted to a central storage site for post-incident examination. The required communication links and archiving of the video data are still expensive and this setup excludes preemptive actions to respond to imminent threats. An effective way to overcome these limitations is to build a smart camera that analyzes the data on-site, close to the sensor, and transmits alerts when relevant video sequences are detected.

Deep neural networks (DNNs) have come to outperform humans in visual classifications tasks and are also performing exceptionally well on other computer vision tasks. The concept of DNNs and Convolutional Networks (ConvNets) can easily be extended to make use of higher-dimensional input data such as multispectral data. We explore this opportunity in terms of achievable accuracy and required computational effort.

To analyze the precision of DNNs for scene labeling in an urban surveillance scenario we have created a dataset with 8 classes obtained in a field experiment. We combine an RGB camera with a 25-channel VIS-NIR snapshot sensor to assess the potential of multispectral image data for target classification. We evaluate several new DNNs, showing that the spectral information fused together with the RGB frames can be used to improve the accuracy of the system or to achieve similar accuracy with a 3x smaller computation effort. We achieve a very high per-pixel accuracy of 99.1%. Even for scarcely occurring, but particularly interesting classes, such as cars, 75% of the pixels are labeled correctly with errors occurring only around the border of the objects. This high accuracy was obtained with a training set of only 30 labeled images, paving the way for fast adaptation to various application scenarios.

## 1. INTRODUCTION

Video analysis is widely used for enhanced surveillance and inspection applications in many commercial and industrial products. Video analysis is based on algorithms that process the images acquired by a camera to extract features and meaning to automatically detect significant events. During the last 20 years many algorithms have been proposed to achieve the best performance in video analysis using several approaches (e.g. Support Vector Machines, Hidden Markov Models among others). Today, novel algorithms based on deep neural networks (DNNs) are overcoming the performance of the previous algorithms and coming close to or exceeding the accuracy of humans. Moreover, DNNs are not only achieving high performance in whole image classification but also in parts of them, i.e. in object detection and in general object/scene semantic segmentation.

---

[*] Corresponding author. E-mail: cavigelli@iis.ee.ethz.ch

Multispectral images are widely used in remote sensing, geophysical monitoring, astronomy, industrial process monitoring and target detection[1–6] because of their ability to capture more information about the material properties, facilitating the analysis of the image data[7] by seeing more than normal RGB cameras or the human eye. However, such cameras with more than just a few channels are still relatively rare, and a large share of them is likely used for defense applications in a stationary setup or mounted on vehicles due to their size, weight, and procurement cost. Over the last two years, a new type of fully-integrated multispectral CMOS sensor has become available, shrinking the device size and weight to the one of normal RGB cameras while at the same time making them much more affordable. The application of DNN-based embedded visual analytics to multispectral images is a scarcely explored area, with significant potential for improved robustness and accuracy.

DNNs are also known to be computationally expensive. A popular way to provide this computation power needed to process images and video, is to use a large number of expensive servers in a datacenter, which are connected to the camera via Ethernet or other fast communication interfaces. This is, for example, the approach used by Google and Facebook to process images and video from users distributed in the world. However, the constantly increasing number of the cameras sending data to servers and supercomputers, is posing the problem of the huge amount of data being transferred and the computational power needed. Moreover, this approach significantly increases the time required to get a meaning from the processing data, limiting the possibility to have real-time processing. In contrast with this approach, an emerging solution is to process the data close to the sensors with embedded processors. This brings several benefits, such as reducing the amount of data needed to be transferred, improving the response time to detect dangerous situations and not requiring any fast external connectivity or huge storage capacity to work properly. Due to these properties, cameras that embed computational resources on board, well-known as smart cameras, are becoming more and more popular and are used in emerging surveillance applications. Although today embedded processors have very high computational power, pushed up by advances in mobile phones and portable computers, they are still orders of magnitude less powerful than supercomputers and workstations. For this reason, video processing in embedded processors remains a very challenging task, especially when they implement highly accurate DNNs. To run these algorithms on an embedded platform, a combination of optimized DNNs and a highly efficient implementation is needed[8–15].

In this paper, we analyze the potential use of multispectral sensors and embeddable DNNs for automated analysis of video surveillance data in a "smart multispectral camera" system. We create a dataset using a 2k RGB camera combined with a multispectral imaging device providing 25 equally-spaced spectral bands in the range of 600 to 975 nm from an urban surveillance perspective. With the obtained data we evaluate the accuracy achievable with several different DNNs analyzing the data automatically, labeling each pixel with one of 8 classes.

The rest of the paper is organized as follows: In Section 2 we list some related work before explaining how the dataset was collected in Section 3. We present three different type of DNNs in Section 4, which we then evaluate and test in Section 5. Section 6 concludes the paper.

## 2. RELATED WORK

The ground-breaking performance of deep learning and convolutional neural networks (ConvNets) in particular is undisputable nowadays. ConvNets have been shown to outperform traditional computer vision approaches by a large margin in many applications areas and they have even proven their beyond-human performance on visual tasks such as image classification. In this paper, we are focusing on scene labeling, sometimes referred to as semantic segmentation, for which ConvNets are showing similarly outstanding results[8,16–18].

Hyper- and multispectral data and images obtained with very specific spectral filters have been successfully used for industrial computer vision (quality control, …) and remote sensing for some time. However, multispectral sensors have been very expensive, bulky, and often required a non-trivial synchronization system. With the recent appearance of single-chip multispectral snapshot sensor, these have become much more comparable with industrial RGB image sensors. Alongside this, new analysis tools have become available, some of them freely like Scyven[19].

Existing work using ConvNets to analyze multispectral image data is limited to different application areas and often very few spectral channels. In one work[20], the authors combine RGB images with a single thermal LWIR channel to detect (but not segment) pedestrians from a car's perspective, collecting a dataset and using traditional HOG features. The authors of two other works[21,22] report on using ConvNets to classify aerial scenes from the UCMerced Land-use dataset[23] (32 classes, 100 images with 256x256 pixel each), the RS19 dataset (1005 images from Google Earth with 600x600 pixel with 19 classes) and the Brazilian Coffee Scenes dataset[24] (2 classes, SPOT satellite images, 36577 images with 64x64 pixels each), of which the last includes multispectral images with 3 channels: red, green and NIR.

Scene labeling has always been a computationally expensive task, requiring powerful GPUs to be able to process just a few low-resolution images each second using ConvNets[8,25,26] and often several minutes per frame with traditional computer vision methods[27–30] to obtain decent quality results. This already assumes the use of optimized software implementations, and currently only specialized hardware implementations using FPGAs or even ASICs can provide reasonable throughput and accuracy within the power limits of an embedded platform[8,9,31,32].

## 3. DATASET COLLECTION

There is only limited related work using many-channel multispectral information to perform scene labeling, and none for urban surveillance scenarios. Furthermore, the types of sensors used in related work strongly focused on using beam splitters and dedicated imaging sensors for each channel instead of a multispectral mosaic sensor[20]. Before being able to perform any evaluations towards answering the question of whether multispectral data can improve scene labeling results or simplify the processing pipeline to obtain good results, we need to create a dataset. We combine a lower resolution multispectral 25-channel mosaic VISNIR sensor with a high resolution RGB camera, which could be integrated with an embedded processor such as the Tegra K1 to build a smart camera able to process data on-site. In this section, we explain how we collected this dataset. We identify the specific cameras used, explain how the data of the two sensors has been merged, and how the ground truth labeling has been created.

### 3.1 Image sensors

We have collected a dataset using two sensors, a high-resolution RGB sensor to capture shapes accurately and a lower resolution multispectral sensor to obtain additional information about the materials. For the RGB images we have used high-resolution camera from Point Grey, the Flea3 FL3-U3-32S2C-CS, built around the Sony IMX036 1/2.8" CMOS sensor providing $2080 \times 1552$ pixel images at 60 frame/s over USB 3. We equipped this camera with a Fujinon YV2.8x2.8SA-2 lens with a variable focal length of 2.8-8 mm.

The multispectral images have been acquired using the Ximea xiSpec MQ022HG-IM-SM5X5-NIR camera, which features a 2/3" CMOS snapshot mosaic sensor by IMEC. This sensor is based on a monochrome CMOSIS CMV2000 device with $2048 \times 1088$ pixels with an additional interference filter-based $5 \times 5$ mosaic to obtain different spectral channels. It provides 25 equally-spaced spectral bands with center frequencies in the range of 600-975 nm and can stream multispectral cubes at up to 170 frame/s over USB 3. We used the Tamron 22HA lens with a focal length of 6.5 mm for this device, combined with a Schneider FIL LP565 long-pass filter to suppress light with half the wavelength of the individual resonance filters creating the mosaic on the sensor. The selection of these lenses was strongly influenced by the desire to capture a surveillance camera view of the scenery[33].

The RGB camera measures 3/3/4.5 cm with a weight of 35 g and the lens is 5 cm long, has a diameter of 5.5 cm and weights 50 g. The multispectral camera weighs 32 g and fits into a 2.7/2.7/3 cm housing. The lens adds another 60 g, is 3.7 cm long and has a 3.7 cm diameter. The two devices use 3 W and 1.6 W, respectively.

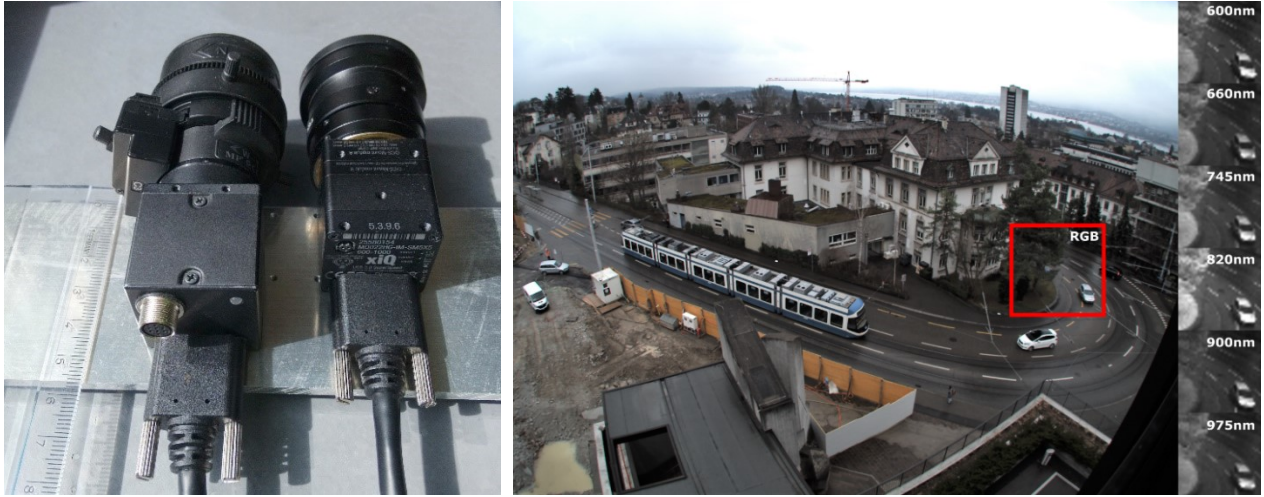

Figure 1. The two cameras fixed on a mounting plate (left) and a sample image of the dataset and cut-outs of 6 from the total of 25 channels of multispectral data (right).

### 3.2 Image alignment

To create a rectified setup, we have fixed both cameras on a mounting plate and adjusted the RGB camera lens' focal length to best match the field-of-view size of the multispectral camera.

In a first step the mosaic multispectral image is converted from its 2D data layout to a multispectral cube of 217 × 409 pixels and 25 channels. When overlaying the images some distortion differences become visible. To correct this, we infer a geometric transformation using the local weighted mean transform (LWMT) with the 12 closest points used to deduce a 2$^{nd}$ degree polynomial transformation for each control point pair based on a total of 33 correspondence pairs scattered all across the image. We use this transform to warp the multispectral image cube to the RGB image using bicubic interpolation, aligning the pixels of the two image sources, such that they can be stacked to a 28 channel image. Finally, the resulting image cube is cropped to 1082 × 1942 pixel, such that only areas where data from both sources is available remain as illustrated in Figure 2.

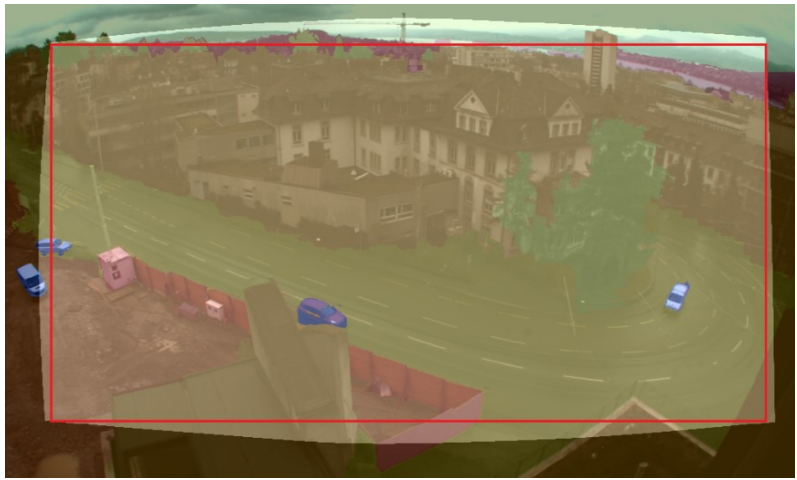

Figure 2. A typical ground-truth labeled RGB image with an overlay showing the area covered by the warped multispectral image and with the cropped area used for our dataset.

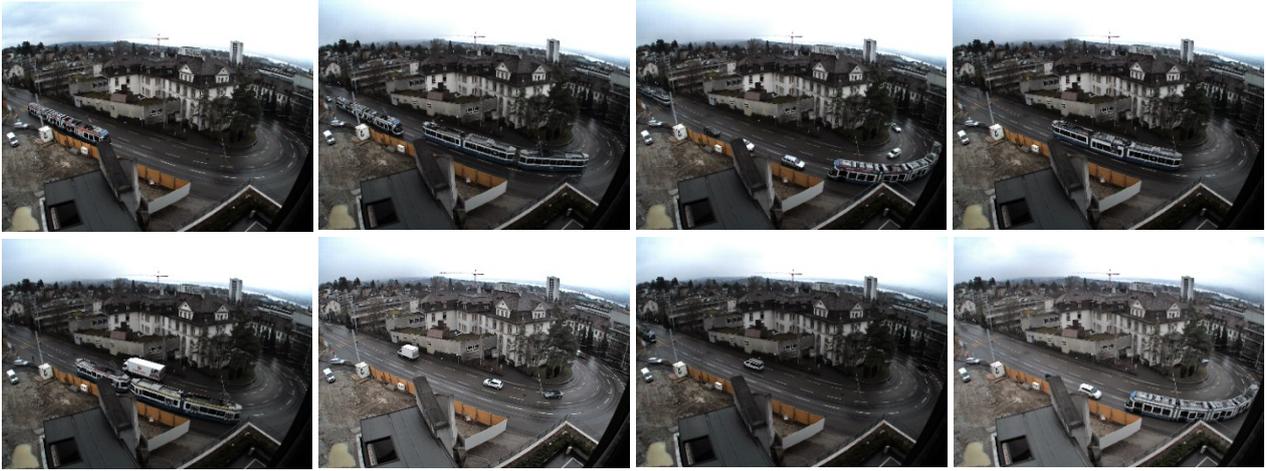

Figure 3. Sample images from the acquired dataset.

With the above mentioned procedure, we do not perform any debayering/demosaicking, for which a large variety of algorithms exist to make the visual perception of RGB images as pleasing as possible. Many of them cannot easily be adapted to non-RGB data, and the most straight-forward one would be to use bilinear interpolation for this as well. We have decided not to do so, because such an interpolation step can also be represented in the first convolution layer of a ConvNet, such that doing this explicitly before would primarily add to the overall computational effort without much benefit. The first convolution layer can also compensate for varying sensitivity of the individual spectral bands.

### 3.3 Data labeling

We have collected 40 images from the same street surveillance perspective. Based on the RGB image, we have labeled each pixel with one of 8 classes: *car/truck, sky, building, road/gravel, tree/shrubbery, tram, water, distant background*. For the evaluation, we have randomly partitioned the dataset into 30 training images and 10 test images. Some sample images are shown in Figure 3.

In order to facilitate the creation of the ground truth, we have developed a program to assist in labeling the dataset shown in Figure 5. Instead of assigning a class to each pixel individually, we segment the image into superpixels using the SLIC algorithm[34] and label these. SLIC clusters the pixels based on a combination of the photo distance and the L2 distance in the image plane to create an oversegmentation of the image. The label is then assigned to each pixel within the superpixel. It has proven useful to start with large superpixels and further improve with a finer-grained segmentation. The manual labeling process was further sped up by making use of the static background, taking the labeling of one image as a starting point for the next one.

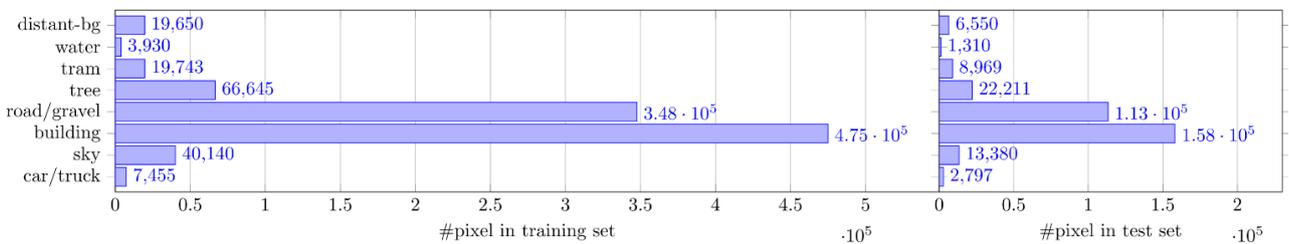

Figure 4. Class distribution of the training and test dataset.

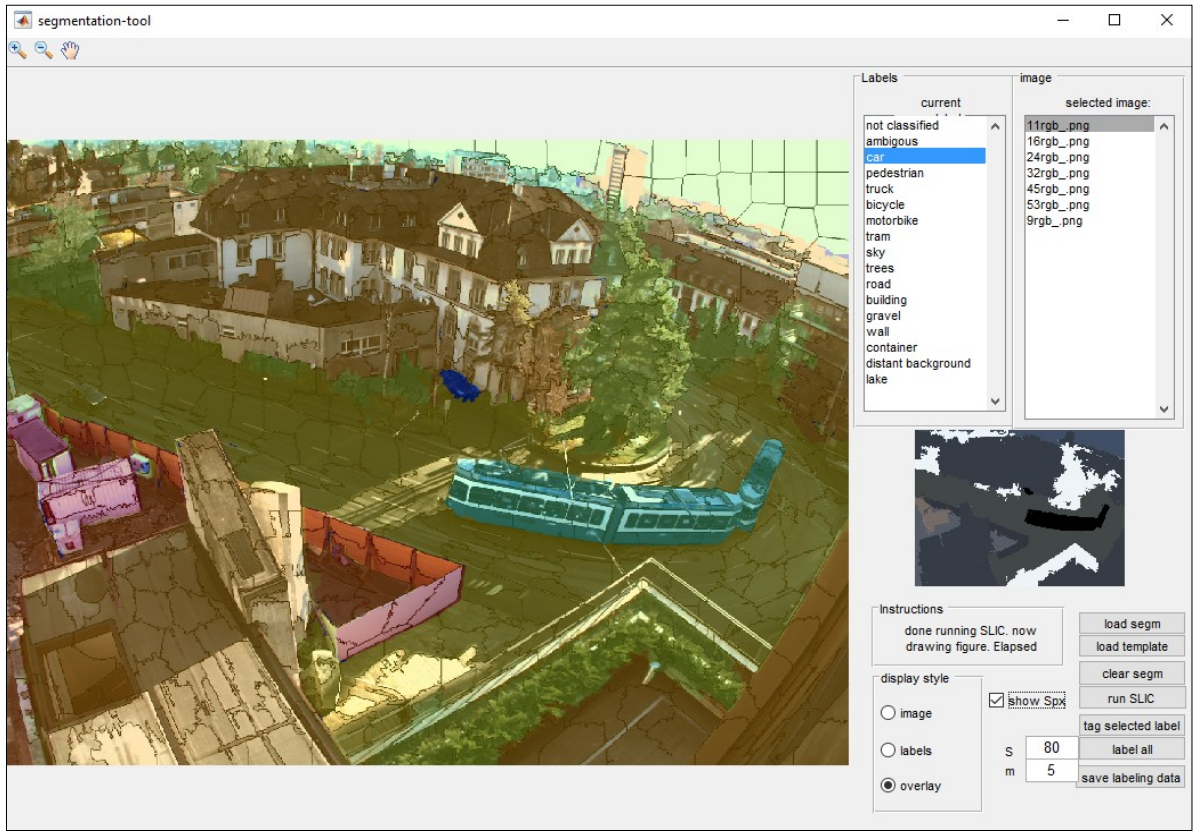

Figure 5. The scene labeling tool with the labeling overlaid on the RGB image and superpixel boundaries in shown in black.

The manual labelling of a dataset is not straight-forward. We have assigned the above labels as accurately as possible, but did not include a *hard-to-classify*, *unclassified*, or *ambiguous* class. This means that if there are pedestrians (a non-existing class) which cover only a few pixels, they have been classified like their surroundings pixels. Also the distinction between *distant background* and *buildings*, *trees* is based on being clearly able to distinguish them at the given resolution, which might vary based on personal perception. Furthermore, the dataset contains trees in front of buildings and the road, leaving some gaps through which the background is visible. We have labeled the entire area covered by the foreground object with its class, only labeling gaps through which the background is visible, if they cover several pixels. The class distribution is very uneven, such that what might be the most interesting classes, *car/truck* and *tram*, make up only a small share of the total number of labeled pixels in the dataset (cf. Figure 4).

## 4. NEURAL NETWORK ARCHITECTURES

In this section, we present three types of neural networks, starting with a per-pixel classification with a normal multi-layer neural network to explore what is possible with a relatively simple classifier. We then move on to present our own proposed ConvNets, exploring the improvement that can be obtained based on the shape and texture of objects in the image. We approach this by adapting a known scene labeling ConvNet targeted at a different, RGB-only dataset and further explore the use of ConvNets based on the concepts of the current state-of-the-art in image recognition and adapt them to our application.

### 4.1 Per-pixel classification

Spectral information has long and successfully been used for material classification. This type of analysis is done on a per-pixel basis, independent of neighboring values. We perform such a material-based classification using a 5-layer neural network evaluated for each pixel individually to analyze whether the additional multispectral channels can improve segmentation results in this setting. This should provide a data point in the corner of fast and very energy-efficient analysis at lower accuracy than the more complex ConvNets.

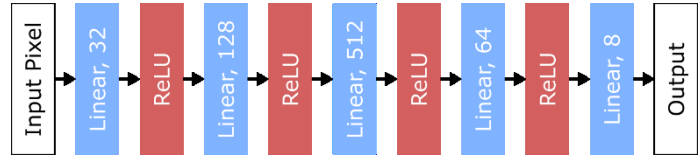

Figure 6. The pixel-wise classification neural network.

We classify each pixel individually by its 3 or 28 channels for the RGB-only and multispectral + RGB image, respectively. We train a 5-layer neural network with 32, 128, 512, 64, and 10 output channels for each layer. As a non-linearity between the layers we use the ReLU activation function, preceded by batch normalization to aid with speedy training[35,36]. As will be explained in the next sections, the other neural networks include two pooling layers, each with a subsampling factor of 2 in both directions. In order to be able to compare these networks better, we subsample the input image before applying pixel-wise classification.

## 4.2 Convolutional network targeting the Stanford backgrounds dataset

Looking only at individual pixels of an image is not optimal, if we desire to obtain a high-quality semantic segmentation of the scene. Based on this input only, it would also be tremendously difficult for humans to solve such a task. A key ingredient to a high-quality segmentation is the recognition of the shape of objects, their texture, and their contextual relation. With a per-pixel analysis there is only a little information about the texture, only the pixel's color, and very little contextual information given by the dataset acquisition setup and its class distribution.

By using a ConvNet we can improve by always taking a spatially local neighborhood into consideration, building a hierarchy of increasingly abstract representations to capture this information, such as that a car has wheels with dark tires and occurs on a road. This is done by the typical stacking of sequences of a convolutional, an activation and a pooling layer.

To get a performance baseline, we used the structure of the network presented in [8], which was optimized for the Stanford Backgrounds dataset. This dataset contains 715 images of street view scenes with 8 classes and a resolution of $320 \times 240$. The used ConvNet is shown in Figure 7, using $2 \times 2$ max-pooling and ReLU activations.

This ConvNet comes in two flavors: single-scale and multi-scale. Applying a feature extractor, such as a ConvNet, on scaled versions of the image to build a limited invariance of the features to the size of objects is a wide-spread concept in computer vision. In order to be able to train the ConvNet end-to-end, the feature extraction part is applied to the images scaled down by a factor of 1, 2, and 4. The resulting feature map is bilinearly interpolated such that all 3 feature maps are of the same size, before they are stack on top of each other and per-pixel classification is applied. The three feature extraction branches for the different scales share the same parameters (weights, biases) and the gradients during backpropagation are applied to all the branches.

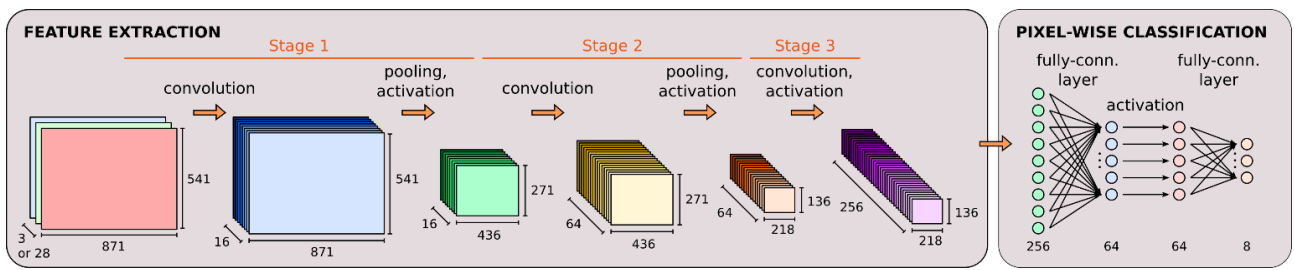

Figure 7. The scene labeling ConvNet based on [8] used to obtain a performance baseline.

Differently from the original implementation, we apply the convolutional layers with zero-padding of their input in order to not loose pixels at the borders of the image when applying the convolutions also in the single-scale configuration. In this way, we obtain a borderless result, since the multi-scale configuration would otherwise imply very wide borders and a full output labeling map is desirable in any case. In order to reduce the evaluation time and memory requirements, we are training the ConvNet with a scaled-down input image of $541 \times 971$ pixel. With this network, the output resolution is smaller by a factor of 4 in each direction due to the max-pooling layers. In order to process not only the RGB data, but also the multispectral information, we simply increase the number of input layers of the first convolutional layer to 28. Contrary to the original network, it has proven to be more effective not to include any preprocessing steps such as contrast normalization.

### 4.3 ResNet-inspired convolutional networks

The network presented in the previous section has been optimized for a different dataset and newer types of networks which are showing better accuracy in image recognition tasks have become available. Deep residual networks (ResNets) are the current state-of-the-art method, having a sequence of very small filters ($3 \times 3$) and every few layers a bypass path to be able to train deeper layers of the network as well[18]. We take this concept to build our own ConvNets and to optimize them for our application. We have constructed several ConvNets of varying depth and different numbers of feature maps, and analyze the effect of multi-scale features and the multispectral data. We have kept the two most interesting results which are at the Pareto-optimal front in terms of error rate and computational effort.

### 4.4 Training the convolutional networks

We use the Torch framework for training the ConvNets. Optimization of the trained parameters is done using the ADAM algorithm[37] and with batch normalization layers in front of the activation functions and no dropout[36]. We apply an equal-weight multi-class margin loss function, $\mathcal{L}(\mathbf{x}, t) = \sum_c \max(0, 1 - x_t - x_c)$ with the the target class index and x the output of the ConvNet. This optimizes the parameters of the network to maximize the distance of the training samples in the output space of the ConvNet, similar to the linear SVM objective[38]. All of the networks used here have an output with lower resolution due to the pooling layers. We train the network with an accordingly down-sampled ground truth labeling.

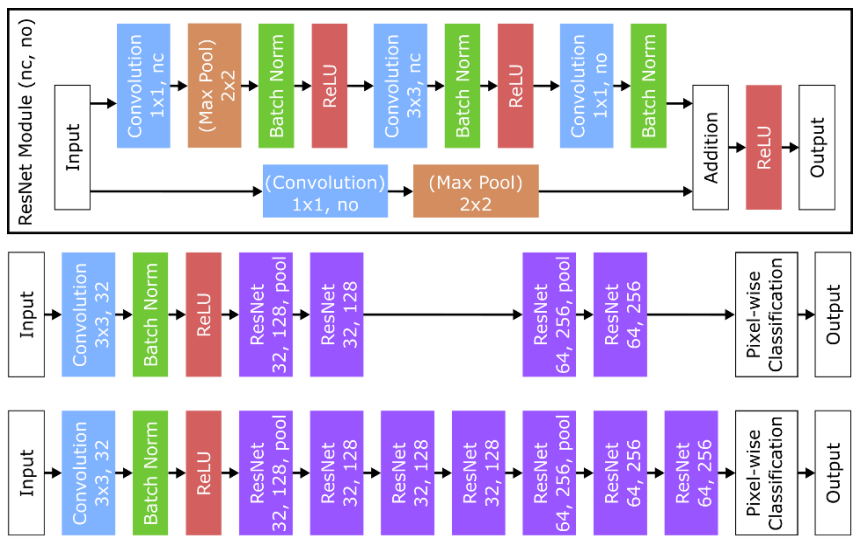

Figure 8. On the top, the ResNet module is shown which serves as a basis for the two ResNet-inspired ConvNets. The max-pooling operations are only applied if mentioned and the convolution module in the lower path is only inserted if the number of feature maps differs between the input and the output. The two networks evaluated are shown in the center and at the bottom of the image. The pixel-wise classification is the same as shown in Figure 7.

## 5. RESULTS & DISCUSSION

We have evaluated the aforementioned neural networks to determine the per-pixel error rate as well as the computational effort, the two most important criteria for a smart camera. Performing per-pixel classification (Network A) using the RGB and the multispectral data, we obtained an error rate of 9.4% on the test set. For comparison, the ConvNet described in Section 4.2 (Network B) yielded an error rate of 0.9% and the two ResNet-like ConvNets achieved error rates of 1.3% and 1.1% for the deeper and the shallower one (Network C1 and C2), respectively. Several additional neural networks based on the concepts shown in Section 4 have been evaluated, but we report only those on the Pareto-optimal front in terms of error rate and computation effort.

One of the major goals in our experiments was to find whether a multispectral camera would add relevant new information to improve the classification accuracy. We have thus trained the best performing Network B only on RGB data as well, suffering a noticeable degradation with an increase of 0.4% in the error rate. This clearly shows that the additionally gathered information aids in improving the quality of the results. It also more generally shows that it is indeed possible to train a ConvNet for this task using a relatively small dataset.

The variations between the analyzed neural networks in terms of error might seem small and a slightly less accurate network acceptable (cf. Figure 9). However, with a dataset as unbalanced as the one obtained for our application scenario, this can have a strongly leveraged impact on the error rate of uncommon classes. The confusion matrices in Figure 10 visualize this. The difference of 0.4% on the per-pixel error rate results in a drop of 18% in the classification of *car/truck* pixels. However, an error rate of e.g. 10% in case of the *tram* does not mean that objects are missed, but rather that the classification around their borders is somewhat fuzzy. It is important to note, that every car, truck or tram in the test set has always been partially recognized.

On an embedded platform with hard power constraints, the computation effort can limit the choice of neural networks deployable. Thus, it can become acceptable to have some accuracy losses and use the networks C1 or C2. The Network C1 has the same error rate as the RGB-only Network A with a 3x smaller computational burden, that makes it interesting to add such a multispectral sensor to save power on a system level. The Nvidia Tegra K1 platform with a two-core ARM processor and a small GPU is specifically targeting such embedded computer vision applications and is able to perform about 96 GOp/s, consuming 10 Watts [8]. This means that we can either process about 1 frame/s with Network A or 3 frame/s with Network C1, achieving the same accuracy as Network A if it was evaluated without the multispectral data.

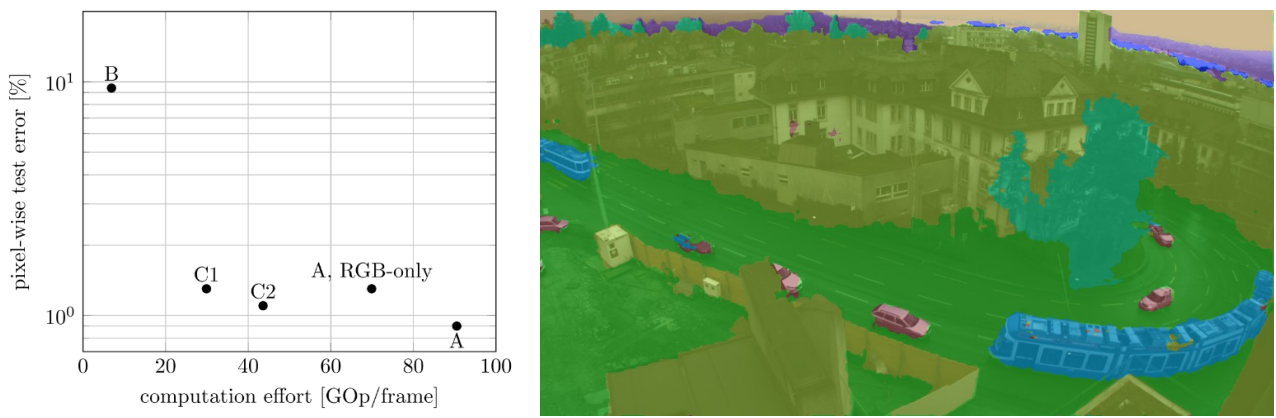

Figure 9. Pixel-wise error rate of the evaluated networks on the test set and their computation effort (left) and scene labeling output of a test image processed by Network A.

| | car/truck | sky | building | road/gravel | tree | tram | water | distant-bg |
|---|---|---|---|---|---|---|---|---|
| car/truck | 0.57 | 0.03 | 0.08 | 0.26 | 0.00 | 0.06 | 0.00 | 0.00 |
| sky | 0.00 | 0.99 | 0.00 | 0.00 | 0.00 | 0.00 | 0.00 | 0.00 |
| building | 0.00 | 0.00 | 1.00 | 0.00 | 0.00 | 0.00 | 0.00 | 0.00 |
| road/gravel | 0.00 | 0.00 | 0.01 | 0.99 | 0.00 | 0.00 | 0.00 | 0.00 |
| tree | 0.00 | 0.00 | 0.01 | 0.00 | 0.99 | 0.00 | 0.00 | 0.00 |
| tram | 0.05 | 0.00 | 0.03 | 0.06 | 0.00 | 0.86 | 0.00 | 0.00 |
| water | 0.00 | 0.03 | 0.06 | 0.00 | 0.00 | 0.00 | 0.89 | 0.02 |
| distant-bg | 0.00 | 0.02 | 0.01 | 0.00 | 0.00 | 0.00 | 0.00 | 0.96 |

Network A, RGB-only. Pixel-wise error rate: 1.3%

| | car/truck | sky | building | road/gravel | tree | tram | water | distant-bg |
|---|---|---|---|---|---|---|---|---|
| car/truck | 0.75 | 0.00 | 0.02 | 0.15 | 0.00 | 0.08 | 0.00 | 0.00 |
| sky | 0.00 | 1.00 | 0.00 | 0.00 | 0.00 | 0.00 | 0.00 | 0.00 |
| building | 0.00 | 0.00 | 1.00 | 0.00 | 0.00 | 0.00 | 0.00 | 0.00 |
| road/gravel | 0.00 | 0.00 | 0.00 | 0.99 | 0.00 | 0.00 | 0.00 | 0.00 |
| tree | 0.00 | 0.00 | 0.00 | 0.00 | 1.00 | 0.00 | 0.00 | 0.00 |
| tram | 0.03 | 0.00 | 0.01 | 0.06 | 0.00 | 0.90 | 0.00 | 0.00 |
| water | 0.00 | 0.00 | 0.00 | 0.00 | 0.00 | 0.00 | 1.00 | 0.00 |
| distant-bg | 0.00 | 0.00 | 0.00 | 0.00 | 0.00 | 0.00 | 0.00 | 1.00 |

Network A. Pixel-wise error rate: 0.9%

Figure 10. Confusion matrices of Network A using the RGB data solely (left) and combining it with the multispectral data (right).

## 6. CONCLUSIONS

In this work we evaluated the benefits of combining an RGB camera with a multispectral camera in an embedded smart camera. We have collected a dataset for scene labeling from an urban surveillance perspective including a multispectral camera. We have presented novel ConvNets for scene labeling using this additional data. We showed that even with a very limited amount of labeled data, highly accurate convolutional networks can be trained, making them an interesting option even for rapid deployment in new surroundings. We have further report how multispectral data can be used to improve the accuracy or, alternatively, to reduce the computational effort by 3x effectively increasing the overall energy efficiency and pushing real-time processing closer the range of what is possible on embedded processing platforms.

## ACKNOWLEDGEMENTS

This work was funded by armasuisse Science & Technology.

## REFERENCES

[1] Gross W., Boehler J., Schilling, H., Middelmann W., Weyermann J., Wellig P., Oechslin R C., Kneubuehler M., "Assessment of target detection limits in hyperspectral data," Proc. SPIE Secur. + Def. **9653** (2015).

[2] Bioucas-dias, J. M., Plaza, A., Camps-valls, G., Scheunders, P., Nasrabadi, N. M.., Chanussot, J., "Hyperspectral Remote Sensing Data Analysis and Future Challenges," IEEE Geosci. Remote Sens. Mag.(June), 6–36 (2013).

[3] Güneralp, I., Filippi, A. M.., Randall, J., "Estimation of floodplain aboveground biomass using multispectral remote sensing and nonparametric modeling," Int. J. Appl. Earth Obs. Geoinf. **33**(1), 119–126 (2014).

[4] Qin, J., Chao, K., Kim, M. S., Lu, R.., Burks, T. F., "Hyperspectral and multispectral imaging for evaluating food safety and quality," J. Food Eng. **118**(2), 157–171, Elsevier Ltd (2013).

[5] Dissing, B. S., Papadopoulou, O. S., Tassou, C., Ersboll, B. K., Carstensen, J. M., Panagou, E. Z.., Nychas, G. J., "Using Multispectral Imaging for Spoilage Detection of Pork Meat," Food Bioprocess Technol. **6**(9), 2268–2279 (2013).

[6] van der Meer, F. D., van der Werff, H. M. A., van Ruitenbeek, F. J. A., Hecker, C. A., Bakker, W. H., Noomen, M. F., van der Meijde, M., Carranza, E. J. M., de Smeth, J. B., et al., "Multi- and hyperspectral geologic remote sensing: A review," Int. J. Appl. Earth Obs. Geoinf. **14**(1), 112–128, Elsevier B.V. (2012).

[7] Quesada-Barriuso, P., Argüello, F.., Heras, D. B., "Efficient segmentation of hyperspectral images on commodity GPUs," Front. Artif. Intell. Appl. **243**, 2130–2139 (2012).

[8] Cavigelli, L., Magno, M.., Benini, L., "Accelerating Real-Time Embedded Scene Labeling with Convolutional Networks," Proc. ACM/IEEE Des. Autom. Conf. (2015).

[9] Andri, R., Cavigelli, L., Rossi, D.., Benini, L., "YodaNN: An Ultra-Low Power Convolutional Neural Network Accelerator Based on Binary Weights," arXiv:1606.05487 (2016).

[10] Rastegari, M., Ordonez, V., Redmon, J.., Farhadi, A., "XNOR-Net: ImageNet Classification Using Binary Convolutional Neural Networks," arXiv:1603.05279 (2016).

[11] Paszke, A., Chaurasia, A., Kim, S.., Culurciello, E., "ENet: A Deep Neural Network Architecture for Real-Time Semantic Segmentation," arXic:1609.02147 (2016).

[12] Ovtcharov, K., Ruwase, O., Kim, J., Fowers, J., Strauss, K.., Chung, E. S., "Accelerating Deep Convolutional Neural Networks Using Specialized Hardware" (2015).

[13] Chen, T., Du, Z., Sun, N., Wang, J., Wu, C., Chen, Y.., Temam, O., "DianNao: A Small-Footprint High-Throughput Accelerator for Ubiquitous Machine-Learning," Proc. ACM Int. Conf. Archit. Support Program. Lang. Oper. Syst., 269–284 (2014).

[14] Farabet, C., Martini, B., Corda, B., Akselrod, P., Culurciello, E.., LeCun, Y., "NeuFlow: A Runtime Reconfigurable Dataflow Processor for Vision," Proc. IEEE Conf. Comput. Vis. Pattern Recognit. Work., 109–116 (2011).

[15] Conti, F.., Benini, L., "A Ultra-Low-Energy Convolution Engine for Fast Brain-Inspired Vision in Multicore Clusters," Proc. IEEE Des. Autom. Test Eur. Conf. (2015).

[16] Farabet, C., Couprie, C., Najman, L.., LeCun, Y., "Scene Parsing with Multiscale Feature Learning, Purity Trees, and Optimal Covers," arXiv:1202.2160 (2012).

[17] Long, J., Shelhamer, E.., Darrell, T., "Fully Convolutional Networks for Semantic Segmentation," Proc. IEEE Conf. Comput. Vis. Pattern Recognit. (2015).

[18] He, K., Zhang, X., Ren, S.., Sun, J., "Deep Residual Learning for Image Recognition," arXiv:1512.03385 (2015).

[19] Habili, N.., Oorloff, J., "Scyllarus: From Research to Commercial Software," Proc. 24th Australas. Softw. Eng. Conf. (2015).

[20] Hwang, S., Park, J., Kim, N., Choi, Y.., Kweon, I. S., "Multispectral pedestrian detection: Benchmark dataset and baseline," Proc. IEEE Comput. Soc. Conf. Comput. Vis. Pattern Recognit., 1037–1045 (2015).

[21] Nogueira, K., Miranda, W. O.., Santos, J. A. Dos., "Improving Spatial Feature Representation from Aerial Scenes by Using Convolutional Networks," Brazilian Symp. Comput. Graph. Image Process. **2015-Octob**, 289–296 (2015).

[22] Nogueira, K., Penatti, O. A. B.., Santos, J. A. dos., "Towards Better Exploiting Convolutional Neural Networks for Remote Sensing Scene Classification," arXiv:1602.01517 (2016).

[23] Yang, Y.., Newsam, S., "Bag-of-visual-words and spatial extensions for land-use classification," Proc. 18th SIGSPATIAL Int. Conf. Adv. Geogr. Inf. Syst. - GIS '10, 270 (2010).

[24] Penatti, A. B., Nogueira, K.., Santos, J. A., "Do Deep Features Generalize from Everyday Objects to Remote Sensing and Aerial Scenes Domains ?," 44–51 (2015).

[25] Farabet, C., Couprie, C., Najman, L.., LeCun, Y., "Learning Hierarchical Features for Scene Labeling," IEEE Trans. Pattern Anal. Mach. Intell. (2013).

[26] Farabet, C., "Towards Real-Time Image Understanding with Convolutional Networks," Université Paris-Est (2014).

[27] Seyedhosseini, M.., Tasdizen, T., "Scene Labeling with Contextual Hierarchical Models," arXiv:1402.0595 (2014).

[28] Kumar, M.., Koller, D., "Efficiently selecting regions for scene understanding," Proc. IEEE Conf. Comput. Vis. Pattern Recognit., 3217–3224 (2010).

[29] Ayvaci, A., Raptis, M.., Soatto, S., "Occlusion Detection and Motion Estimation with Convex Optimization," Adv. Neural Inf. Process. Syst. **2**, 100–108 (2010).

[30] Tighe, J.., Lazebnik, S., "Superparsing: scalable nonparametric image parsing with superpixels," Proc. Eur. Conf. Comput. Vis. (2010).

[31] Cavigelli, L., Gschwend, D., Mayer, C., Willi, S., Muheim, B.., Benini, L., "Origami: A Convolutional Network Accelerator," Proc. ACM Gt. Lakes Symp. VLSI, 199–204, ACM Press (2015).

[32] Cavigelli, L.., Benini, L., "Origami: A 803 GOp/s/W Convolutional Network Accelerator," IEEE Trans. Circuits Syst. Video Technol. (2016).

[33] Kruegle, H., CCTV Surveillance: Video Practices and Technology, Butterworth-Heinemann, Woburn, MA, USA (1995).

[34] Achanta, R., Shaji, A.., Smith, K., "SLIC Superpixels Compared to State-of-the-Art Superpixel Methods," Pattern Anal. … **34**(11), 2274–2281 (2012).

[35] Nair, V.., Hinton, G. E., "Rectified Linear Units Improve Restricted Boltzmann Machines," Proc. 27th Int. Conf. Mach. Learn.(3), 807–814 (2010).

[36] Ioffe, S.., Szegedy, C., "Batch Normalization: Accelerating Deep Network Training by Reducing Internal Covariate Shift," Proc. Int. Conf. Mach. Learn., 448–456 (2015).

[37] Kingma, D.., Ba, J., "Adam: A Method for Stochastic Optimization," Proc. Int. Conf. Learn. Represent. (2015).

[38] Rosasco, L., De Vito, E., Caponnetto, A., Piana, M.., Verri, A., "Are loss functions all the same?," Neural Comput. **16**(5), 1063–1076 (2004).